\documentclass[pmlr,twocolumn]{jmlr}

\usepackage{longtable}
\usepackage{booktabs}
\usepackage{pifont}
\usepackage{enumitem}
\usepackage[load-configurations=version-1]{siunitx}

\theorembodyfont{\upshape}
\theoremheaderfont{\scshape}
\theorempostheader{:}
\theoremsep{\newline}

\jmlrvolume{193}
\firstpageno{1}

\jmlryear{2022}
\jmlrworkshop{Machine Learning for Health (ML4H) 2022}

\title[Adapting Pre-trained Vision Transformers from 2D to 3D Improves Segmentation]{Adapting Pre-trained Vision Transformers from 2D to 3D through Weight Inflation Improves Medical Image Segmentation}

\author{\Name{Yuhui Zhang} \Email{yuhuiz@stanford.edu}\\
\Name{Shih-Cheng Huang} \Email{mschuang@stanford.edu}\\
\Name{Zhengping Zhou} \Email{zpzhou@stanford.edu}\\
\Name{Matthew P. Lungren} \Email{mlungren@stanford.edu}\\
\Name{Serena Yeung} \Email{syyeung@stanford.edu}\\
\addr Stanford University, Stanford, CA 94305, USA}

\begin{document}

\maketitle

\begin{abstract}
Given the prevalence of 3D medical imaging technologies such as MRI and CT that are widely used in diagnosing and treating diverse diseases, 3D segmentation is one of the fundamental tasks of medical image analysis.
Recently, Transformer-based models have started to achieve state-of-the-art performances across many vision tasks, through pre-training on large-scale natural image benchmark datasets.
While works on medical image analysis have also begun to explore Transformer-based models, there is currently no optimal strategy to effectively leverage pre-trained Transformers, primarily due to the difference in dimensionality between 2D natural images and 3D medical images.
Existing solutions either split 3D images into 2D slices and predict each slice independently, thereby losing crucial depth-wise information, or modify the Transformer architecture to support 3D inputs without leveraging pre-trained weights. 
In this work, we use a simple yet effective weight inflation strategy to adapt pre-trained Transformers from 2D to 3D, retaining the benefit of both transfer learning and depth information. 
We further investigate the effectiveness of transfer from different pre-training sources and objectives.
Our approach achieves state-of-the-art performances across a broad range of 3D medical image datasets, and can become a standard strategy easily utilized by all work on Transformer-based models for 3D medical images, to maximize performance.\footnote{Codes are available at \url{https://github.com/yuhui-zh15/TransSeg}.}
\end{abstract}
\begin{keywords}
Medical Image Segmentation, Transfer Learning, CT, MRI.
\end{keywords}

\section{Introduction}

\begin{figure*}[t]
    \centering
    \includegraphics[width=\linewidth]{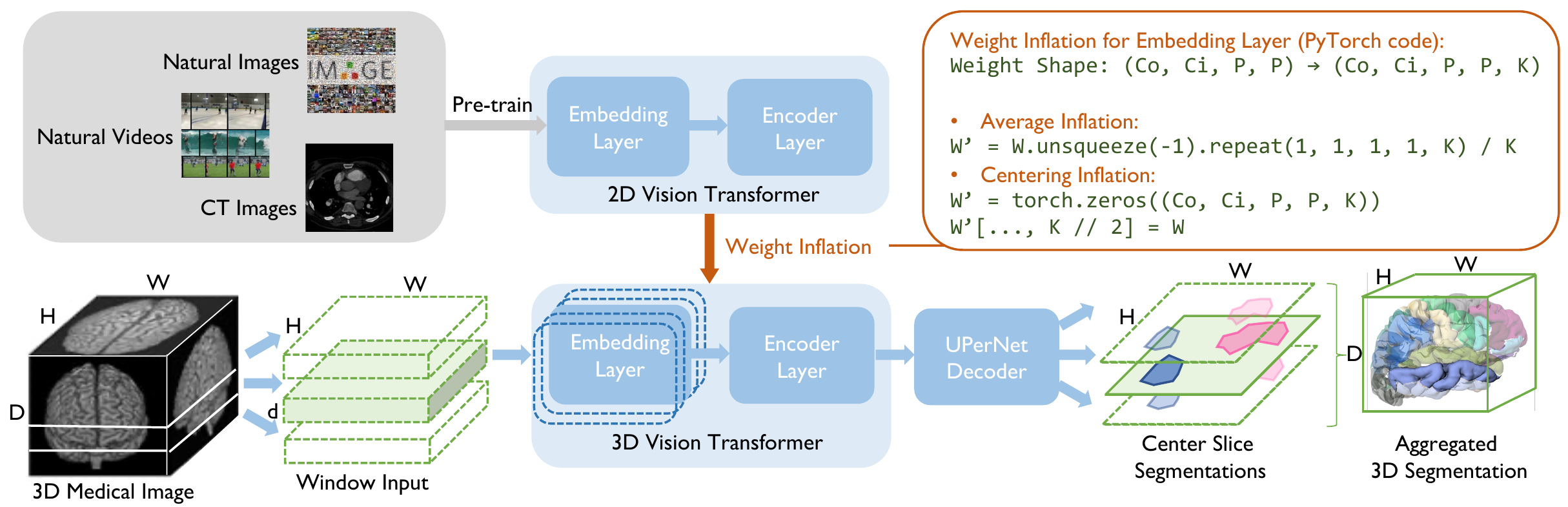}
    \caption{\emph{Approach overview.} Large-scale pre-trained Transformers are used as the encoder in the segmentation model for transfer learning, in which weights are adapted using the inflation strategy to support 3D inputs. Each 3D image is split into windows, which contain a small number of neighbor slices. Each window is fed into the segmentation model and the segmentation of the center slice is predicted. All the predicted slices are aggregated to form the final 3D prediction.}
    \label{fig:method}
\end{figure*}

The increased utilization of medical imaging technologies in recent years is causing a dramatic increase in radiologists' daily workload~\citep{smith2008rising, hendee2010addressing}, leading to longer delays in diagnosis and higher misdiagnosis rates~\citep{alonso2010delay, hendriksen2017clinical}. Computer vision techniques such as 3D segmentation have the potential to alleviate the burden for radiologists by automating or assisting current image interpretation workflow, ultimately improving clinical care and patient outcome~\citep{esteva2021deep, huang2020fusion}. 

Recently, the computer vision community has witnessed a paradigm shift from using CNNs to self-attention based Transformer~\citep{vaswani2017attention} architectures. By pre-training at scale using supervised or self-supervised learning, many vision Transformer models have achieved state-of-the-art performances when fine-tuned for various vision tasks, including image classification, object detection, and semantic segmentation~\citep{caron2021emerging, liu2021video, bao2021beit, carion2020end, dosovitskiy2020image}.

However, while many recent works on medical image analysis have started to explore Transformer-based models, it is still unclear what is the optimal strategy to effectively leverage Transformers for 3D medical image segmentation, primarily due to the difference in dimensionality between 2D natural images and 3D medical images. 
A straightforward approach adopted by most existing works is to split 3D images into 2D slices along the depth axis and independently segment each of them~\citep{chen2021transunet,liu2021swin,huang2021missformer}. However, this solution compromises the depth information, which is crucial for identifying segmentation boundaries.
On the contrary, another line of research modifies the Transformer architecture to support 3D inputs, but the modification makes the pre-trained weights not directly applicable~\citep{xie2021cotr,hatamizadeh2021unetr}. 

In this work, we investigate strategies and best practices for adapting  Transformers for 3D medical images. We first analyze the importance of \emph{transfer learning} and \emph{depth information}, and find that both are critical for segmentation performance, especially transfer learning. The model initialized from pre-trained weights outperforms the same randomly initialized model by 11.18\% on a multi-organ segmentation dataset, and the randomly initialized 3D model outperforms the randomly initialized 2D model by 6.07\%. 
To retain these two advantages, we use a simple yet effective \emph{weight inflation} strategy to adapt pre-trained Transformers from 2D to 3D, which has been a standard approach in video understanding to transfer from models pre-trained on images since it was first proposed in I3D~\citep{carreira2017quo}. After careful ablations of different inflation and transfer settings, our best strategy achieves 1.95\% improvements over baselines with only a 0.75\% increase in computational cost, and outperforms all the state-of-the-art methods.

We further evaluate our method on 11 additional datasets to understand the generalizability of our best practice. Experiments show that our approach consistently benefits from weight inflation and achieves many state-of-the-art performances. 

In summary, the major contribution of our work to the medical image segmentation community is that we show how a simple yet effective weight inflation strategy, which is not currently used by state-of-the-art medical image segmentation methods, can lead to substantial improvements in performance. Moreover, we also performed systematic ablations, which further provide insights into the best strategies to adapt pre-trained Transformers from 2D to 3D through weight inflation. Our method can become a standard strategy utilized by all future works on Transformer-based models for 3D medical images to maximize performance.

Our best practice is summarized in Figure~\ref{fig:method} and below: 
\begin{itemize}
    \item Find a vision Transformer model pre-trained on natural images via a combination of self-supervised and supervised learning. 
    \item Adapt pre-trained weights in the embedding layer using the centering inflation strategy by transferring weights to the center-most slice and initializing all other weights to zero.
    \item Parse 3D input into windows of a few neighbor slices, and segment only the center slice. Aggregate 3D segmentation with predictions from all windows. 
\end{itemize}

\section{Related Works}

\paragraph{Vision Transformers.} 

Originally proposed for machine translation~\citep{vaswani2017attention}, Transformer based architectures have started to flourish in computer vision in recent years. ViT~\citep{dosovitskiy2020image} split images into patches and create patch embeddings as inputs to the Transformer model, which lays the foundation to use Transformer architectures for visual recognition. However, ViT requires more training data to surpass convolutional neural networks. More recent works explored different self-supervised pre-training strategies to reduce the requirement of large-scale labeled datasets. Caron et al.~\citep{caron2021emerging} proposed DINO, a self-distillation contrastive framework. Drawing inspiration from BERT in the NLP field, Bao et al.~\citep{bao2021beit} proposed BEiT, which is pre-trained to predict masked image patches given context. These approaches achieve state-of-the-art performances in image classification and semantic segmentation. 

Several works have explored leveraging Transformers for video inputs by utilizing weights from pre-trained image models~\citep{carreira2017quo}. ViViT~\citep{arnab2021vivit} and Video SwinTransformer~\citep{liu2021video} explored inflation and centering strategies to initialize weights of video Transformers from pre-trained image Transformers. In this work, we systematically investigate the benefit of transfer learning from these Transformer models under significant domain shifts, and compare the effectiveness of transferring from different sources and pre-training objectives. 

\paragraph{Transformers for 3D Medical Image Segmentation.}

Following successful adaptations of Transformers in vision, some works in medical image segmentation also achieved state-of-the-art results by utilizing Transformers. Earlier work such as TransUNet~\citep{chen2021transunet} or CoTr~\citep{xie2021cotr} still relied on convolutional layers for the Transformer model. More recent methods, such as SwinUNet~\citep{cao2021swin} and UNETR~\citep{xie2021cotr}, use a purely Transformer-based encoder. To improve the information flow of the segmentation framework, different mechanisms such as context bridge~\citep{huang2021missformer} or interleaved layers~\citep{gao2021utnet} are designed.

Based on their model input dimension, these prior works can be grouped into two categories: 3D approaches that develop 3D model architectures directly predicting 3D segmentation~\citep{xie2021cotr,hatamizadeh2021unetr}; and 2D approaches that split 3D images into 2D slices and stack 2D predictions to form the 3D segmentation~\citep{chen2021transunet,cao2021swin,huang2021missformer}.

3D segmentation benefits from having more depth-wise contextual information, which is especially important for medical images where awareness of anatomical structures is crucial for identifying segmentation boundaries. However, due to the limited availability of pre-trained 3D Transformers, all prior 3D approaches randomly initialized their models and used no transfer learning. On the contrary, 2D approaches can easily utilize Transformer weights pre-trained on natural images, but lack the spatial context awareness of 3D models. In this work, we attempt to address the shortcomings of existing approaches by combining the benefit of transfer learning of 2D approaches and depth-wise information of 3D approaches.

\section{Method}

In this section, we first introduce semantic segmentation (Sec.~\ref{section:segmentation}) and vision Transformer (Sec.~\ref{section:architecture}), and then present strategies for adapting Transformers pre-trained on 2D image to 3D medical image inputs (Sec.~\ref{section:transfer}).

\subsection{Semantic Segmentation}
\label{section:segmentation}

Semantic segmentation is the task of assigning a class label for each pixel (voxel for 3D) of an input image. 
Most prior works use \emph{encoder-decoder} frameworks for semantic segmentation. 
The encoder progressively extracts higher-level features that capture the input image's global information.
Then the decoder utilizes these features and progressively reconstructs fine-grained pixel/voxel-level predictions.
Therefore, the quality of the features extracted by the encoder directly impacts the segmentation quality. 

CNNs have been the dominant choices of the encoder in semantic segmentation models for years~\citep{long2015fully,ronneberger2015u}. Recently, several studies have reported improvements in performance for both natural and medical image segmentation by using Transformers as encoders~\citep{liu2021swin,bao2021beit,zheng2021rethinking,chen2021transunet,cao2021swin}. In this work, we compare different pre-trained Transformers as encoders and study the most effective way to perform transfer learning for 3D medical images. 

\subsection{Vision Transformer}
\label{section:architecture}

Transformer was first introduced in~\citep{vaswani2017attention} and adapted to visual inputs in~\citep{dosovitskiy2020image}. Unlike CNNs, Transformer uses the self-attention mechanism to aggregate information from the input image, which can be viewed as more expressive convolutions and can capture long-range dependencies~\citep{cordonnier2019relationship}. Empirical results show that vision Transformers achieve state-of-the-art performances across many vision tasks when pre-trained on large-scale natural image datasets~\citep{dosovitskiy2020image,liu2021swin,caron2021emerging,bao2021beit}.

\paragraph{Patch Partitioning and Embedding Layer.} Vision Transformers first use a \emph{patch partitioning and embedding layer} to convert a visual input into patch embeddings to reduce input dimensionality. Specifically, for a 2D image with $H\times W$ pixels, the image is partitioned into multiple smaller patches of $P \times P$ pixels ($P \in \{16, 32\}$ in practice). Similarly, 3D images of size $H\times W\times D$ can be partitioned into patches of $P\times P\times P$ voxels. A linear layer is then applied to each patch's flattened pixel/voxel values to generate input embeddings to the Transformer encoder. By using patch partitioning, the number of inputs $T= \frac{H}{P} \times \frac{W}{P}$ ($T= \frac{H}{P} \times \frac{W}{P} \times \frac{D}{P}$ for 3D) is much smaller than the number of pixels/voxels, which makes the computation tractable for subsequent Transformer encoder layers given the $O(T^2)$ computational cost. Notably, the input patch partitioning and embedding layer is equivalent to strided convolution, where the kernel size and stride size are both equal to $P \times P$ ($P \times P \times P$ for 3D). This is important as we introduce weight inflation strategies to adapt Transformers from 2D to 3D in Sec~\ref{section:transfer}.

\paragraph{Transformer Encoder.} These patch embeddings $H^0 \in \mathbb{R}^{T\times D}$ are then provided as a sequence input into a \emph{Transformer encoder} that generates contextualized representations for each patch $H^L \in \mathbb{R}^{T\times D}$. 
The encoder is constructed with a stack of $L$ multi-head self-attention layers, which facilitates the aggregation of information from all the other input patches.
After information exchange from every layer in a sequential manner $H^{l}=\text{Layer}_l(H^{l-1}) \in \mathbb{R}^{T\times D}$, the output vectors of the last layer $H^L$ are used as contextualized representations for the input patches. We refer to~\citep{rush2018annotated,vaswani2017attention} for a more detailed description of the Transformer model.

\paragraph{Variations of Vision Transformers.} While many vision Transformers' variations exist, they all have similar architectures as described above.
In this work, we compare four variations: DINO~\citep{caron2021emerging}, BEIT~\citep{bao2021beit}, SwinT~\citep{liu2021swin}, and VideoSwinT~\citep{liu2021video}. 
DINO and BEiT have the same architectures as the original vision Transformer~\citep{dosovitskiy2020image} and only differ in the self-supervised pre-training objectives. SwinT proposes shifted window attention and hierarchical feature size shrinking to introduce inductive bias of locality.
VideoSwinT is an adaptation of SwinT for video inputs.
As each variation has many size configurations, we use all the base models with around 85M parameters for fair comparisons.

\begin{table*}[htbp]
\centering
\small
\caption{\emph{Details of the twelve publicly available 3D medical image segmentation datasets we used, including BCV, ACDC, and 10 MSD datasets}. These datasets cover major anatomical structures of the body and different modalities. We use the BCV dataset to develop our methods and compare them with existing methods, and use the remaining eleven datasets to test the generalizability of our best practices. } 
\begin{tabular}{lcccccc}
\toprule
\textbf{Data} & \textbf{Modality} & \textbf{Raw Reso.} & \textbf{Input Reso.} & \textbf{Clip Range} & \textbf{Label} & \textbf{Split} \\
\midrule
BCV & CT & 512$\times$512$\times$126 & 512$\times$512$\times$5 & [-175, 250] & 13 & 18/12/0 \\
ACDC & CT & 220$\times$247$\times$10 & 512$\times$512$\times$5 & [-175, 250] & 3 & 70/10/20 \\
Brain & MRI & 240$\times$240$\times$155  & 240$\times$240$\times$5 & N/A & 3 & 387/72/25 \\
Heart & MRI & 320$\times$320$\times$114 & 320$\times$320$\times$5 & N/A & 1 & 16/3/1 \\
Liver & CT & 512$\times$512$\times$447 & 512$\times$512$\times$5 & [-175, 250] & 2 & 104/20/7  \\
Hippocampus & MRI & 34$\times$49$\times$36 & 240$\times$240$\times$5 & N/A & 2 & 208/39/13  \\
Prostate & MRI & 320$\times$320$\times$19 & 320$\times$320$\times$5 & N/A & 2 & 25/5/2 \\
Lung & CT & 512$\times$512$\times$282 & 512$\times$512$\times$5 & [-175, 250] & 1 & 50/9/4 \\
Pancreas & CT & 512$\times$512$\times$94 & 512$\times$512$\times$5 & [-175, 250] & 2 & 224/42/15 \\
Vessel & CT & 512$\times$512$\times$69 & 512$\times$512$\times$5 & [-175, 250] & 2 & 242/45/16 \\
Spleen & CT & 512$\times$512$\times$88 & 512$\times$512$\times$5 & [-175, 250] & 1 & 32/6/3\\
Colon & CT & 512$\times$512$\times$110 & 512$\times$512$\times$5 & [-175, 250] & 1 & 100/19/7 \\
\bottomrule
\end{tabular}
\label{tab:data-statistics}
\end{table*}

\subsection{Adapting Transformers from 2D to 3D through Weight Inflation}
\label{section:transfer}

Many variations of Transformers that have been pre-trained on existing large-scale datasets are publicly available for transfer learning~\citep{caron2021emerging,bao2021beit,liu2021swin,liu2021video}. However, direct utilization of these Transformers is non-trivial, due to the difference in the dimensionality between 2D natural images and 3D medical images. Most existing solutions either split 3D images into 2D slices and predict each slice independently, thereby losing crucial depth-wise information~\citep{chen2021transunet,cao2021swin}. Other solutions modify the Transformer architecture to support 3D inputs, but the modification makes pre-trained weights not directly applicable~\citep{xie2021cotr,hatamizadeh2021unetr}. 

We combine both advantages inspired by recent works in video understanding. Videos are collections of 2D image frames organized along the temporal axis, similar to medical images where 2D slices are organized along the depth axis. I3D~\citep{carreira2017quo} proposed to transfer CNNs pre-trained on 2D images to 3D video inputs by inflating the convolutional weights along the temporal axis. As the input patch partitioning and embedding layer in Transformer is equivalent to CNN, we adopt the weight inflation strategy to initialize the 3D Transformers.

We compare two inflation strategies: \emph{average inflation} and \emph{centering inflation}. For average inflation, we copy the weights of the input layer $K$ times in the depth axis and divide them by $K$. This setting assumes input slices are similar within a certain range of depths, and the model treats all the input slices equally at the beginning. For centering inflation, we transfer pre-trained weights for the center-most slice and initialize all other weights to zero. For this setting, the model uses only information from the center slice at the beginning and progressively learns to contextualize information from all the neighbor slices. Both strategies keep the mean and variance of the input to Transformer encoder, allowing more effective transfer learning.

Another difference is in the input channel. Transformers pre-trained on colored natural images have three channels. Similarly, We reduce the channel to one by modifying the input layer weight, where we take the sum over the input channel axis. We further use average inflation in the input channel for medical images with more than one input channel such as MRIs. These modifications still keep the mean and variance of the input unchanged for the Transformer encoder. After adapting the weight in the input patch partition and embedding layer, the weights of Transformer encoder layers can be directly initialized from 2D pre-trained Transformers since input shapes are aligned.

\section{Results}

\begin{table*}[t]
\centering
\small
\caption{\emph{Segmentation results on the BCV dataset.} \emph{(a)} Our approach achieves state-of-the-art performances by combining the advantages of transfer learning (T) and depth information (D). DSC is the Dice coefficient averaged on the 8 major organs. \emph{(b)} Transfer effectiveness from models pre-trained with different sources and objectives. NI, NV, and MI are natural images, natural videos, and medical images. SL and SSL are supervised learning and self-supervised learning. }
\label{tab:bcv_seg}
\subtable[
\textbf{Main Result}
]{
\centering
\begin{minipage}{0.37\linewidth}{\begin{center}
\begin{tabular}{lccc}
\toprule
\textbf{Method} & \textbf{T} & \textbf{D} & \textbf{DSC} \\
\midrule
TransUNet~(\citeyear{chen2021transunet}) & \ding{52} & \ding{54} & 84.36 \\ 
SwinUNet$_\downarrow$~(\citeyear{cao2021swin}) & \ding{52} & \ding{54} & 81.12 \\
\midrule
UNETR~(\citeyear{hatamizadeh2021unetr}) & \ding{54} & \ding{52} & 80.07 \\ 
\midrule
Ours w/o T\&D & \ding{54} & \ding{54} & 74.00 \\
Ours w/o T & \ding{54} & \ding{52} & 74.70 \\
Ours w/o D & \ding{52} & \ding{54} & 85.18 \\
Ours & \ding{52} & \ding{52} & \textbf{87.13} \\
 \bottomrule
\end{tabular} 
\end{center}}\end{minipage}
}
\subtable[
\textbf{Transfer Source and Objective}
]{
\begin{minipage}{0.58\linewidth}{\begin{center}
\begin{tabular}{lllc}
\toprule
\textbf{Encoder} & \textbf{Source} & \textbf{Objective} & \textbf{DSC} \\ 
\midrule
BEiT-Random & - & - & 74.70 \\
CT-DINO & MI \tiny{(RSNACT)} & SSL & 82.56 \\
BEiT-SSL~(\citeyear{bao2021beit}) & NI \tiny{(IN22K)} & SSL & 84.41 \\
DINO~(\citeyear{caron2021emerging}) & NI \tiny{(IN1K)} & SSL & 84.60 \\
SwinT~(\citeyear{liu2021swin}) & NI \tiny{(IN22K)} & SL & 85.57 \\
VideoSwinT~(\citeyear{liu2021video}) & NV \tiny{(K600)} & SL & 85.94 \\
BEiT~(\citeyear{bao2021beit}) & NI \tiny{(IN22K)} & SSL+SL & \textbf{87.13} \\ 
\bottomrule
\end{tabular}
\end{center}}\end{minipage}
}
\end{table*}

\begin{table*}[htbp]
\centering
\small
\caption{\emph{Fine-grained segmentation results of different organs on the BCV dataset.}  We report performances on Aorta, Gallbladder, Kidney (Left), Kidney (Right), Liver, Pancreas, Spleen, and Stomach, respectively. The most significant improvements of our method are Gallbladder, Pancreas, and Stomach segmentation, which can be interpreted as the larger the variation between neighbor slices, the more improvement from incorporating depth information through weight inflation. 
}
\begin{tabular}{lcccccccc}
\toprule
\textbf{Method} & \textbf{Aor} & \textbf{Gal} & \textbf{KidL} & \textbf{KidR} & \textbf{Liv} & \textbf{Pan} & \textbf{Spl} & \textbf{Sto} \\
\midrule
TransUNett~(\citeyear{chen2021transunet}) & 90.68 & 71.99 & 86.04 & 83.71 & 95.54 & 73.96 & 88.80 & 84.20 \\
SwinUNet$_\downarrow$~(\citeyear{cao2021swin}) & 87.07 & 70.53 & 84.64 & 82.87 & 94.72 & 63.73 & 90.14 & 75.29 \\
\midrule
UNETR~(\citeyear{hatamizadeh2021unetr}) & 86.30 & 63.79 & 84.94 & 83.93 & 95.94 & 58.01 & 88.95 & 78.76 \\
\midrule
Ours w/o T\&D & 82.71 & 61.34 & 80.73 & 76.25 & 94.37 & 46.52 & 82.62 & 67.46 \\
Ours w/o T & 81.26 & 63.78 & 80.90 & 75.41 & 94.82 & 53.67 & 82.15 & 65.64 \\
Ours w/o D & 90.22 & 67.51 & 93.84 & 91.40 & 96.27 & 66.62 & 94.22 & 81.33 \\
Ours & 90.29 & 75.20 & 93.70 & 91.40 & 96.17 & 71.82 & 93.81 & 84.62 \\
\midrule
\# Voxels & 441936 & 118445 & 887310 & 879180 & 9762041 & 504539 & 1386739 & 2641170 \\
\# Slices & 827 & 131 & 362 & 350 & 631 & 316 & 332 & 421 \\
Slice Variation & 90.33 & 69.84 & 82.80 & 82.57 & 88.93 & 75.91 & 83.37 & 84.94 \\
\bottomrule
\end{tabular}
\label{tab:organ_analysis}
\end{table*}

\subsection{Dataset}
\label{section:dataset}

We use 12 publicly available 3D medical image datasets in this study: Beyond the Cranial Vault (BCV) multi-organ segmentation dataset~\citep{BCV}, Automated Cardiac Diagnosis Challenge (ACDC)~\citep{ACDC}, and 10 datasets from Medical Segmentation Decathlon (MSD)~\citep{antonelli2021medical}.
These datasets cover major anatomical structures of the body and different modalities such as CT and MRI. 
We use the same BCV training and validation set as existing works~\citep{chen2021transunet,cao2021swin} to develop our method and fairly compare it with other methods. We use the remaining 11 datasets to test the generalizability of our best practice.
Details of the datasets such as labels, modality, resolution, data split are available in Table~\ref{tab:data-statistics}. 

\subsection{Main Result}

\begin{figure*}[!tb]
    \centering
    \hspace{3.5em} Input \hspace{3em} Label \hspace{3em} Pred \hspace{1.5em} Pred w/o D \hspace{0.2em} Pred w/o T\&D \\
    \begin{minipage}[htbp]{0.75\linewidth}
        \centering
        \includegraphics[width=\linewidth]{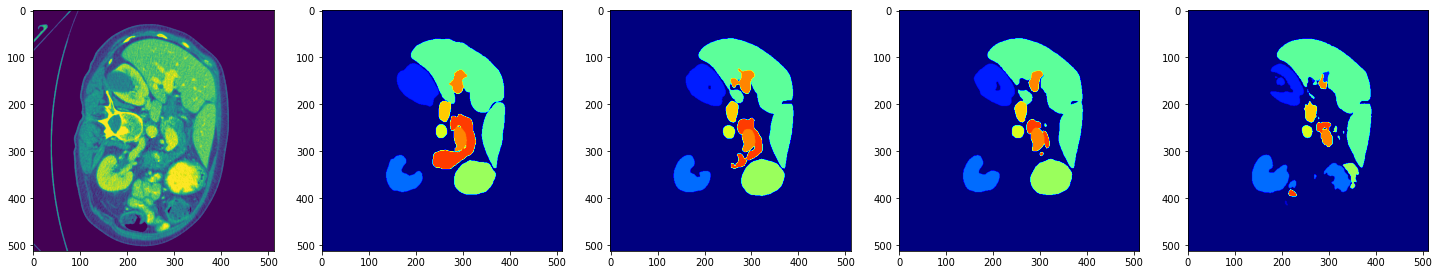}
    \end{minipage} \\
    \begin{minipage}[htbp]{0.75\linewidth}
        \centering
        \includegraphics[width=\linewidth]{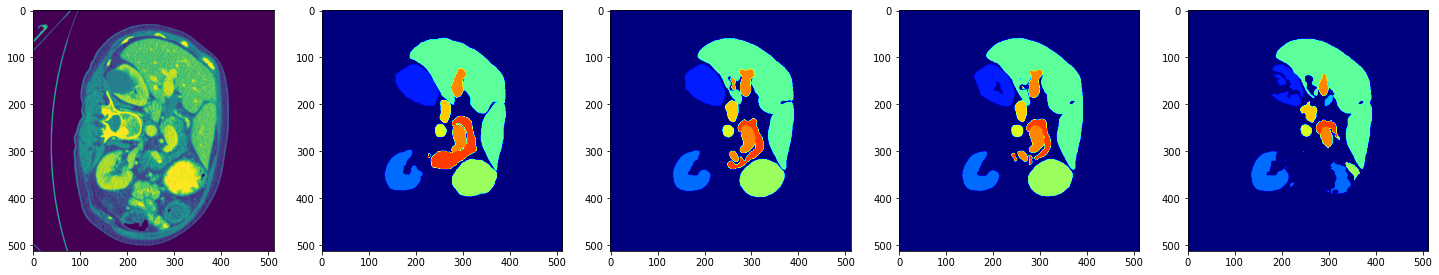}
    \end{minipage} \\
    \begin{minipage}[htbp]{0.75\linewidth}
        \centering
        \includegraphics[width=\linewidth]{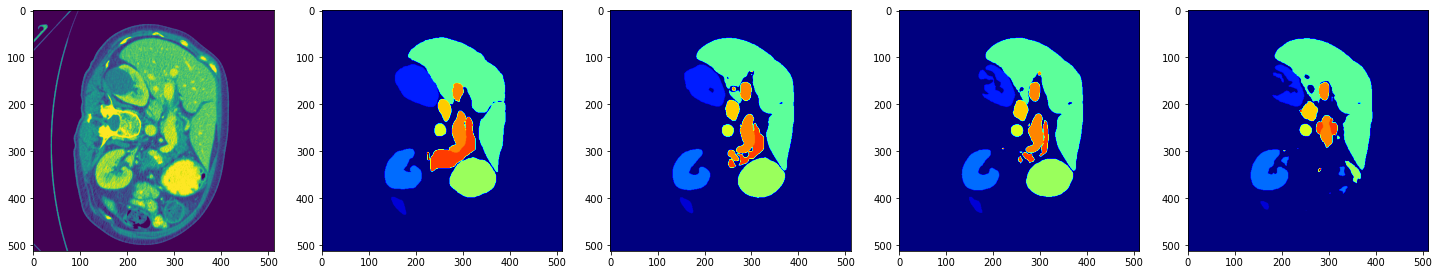}
    \end{minipage} \\
    \begin{minipage}[htbp]{0.75\linewidth}
        \centering
        \includegraphics[width=\linewidth]{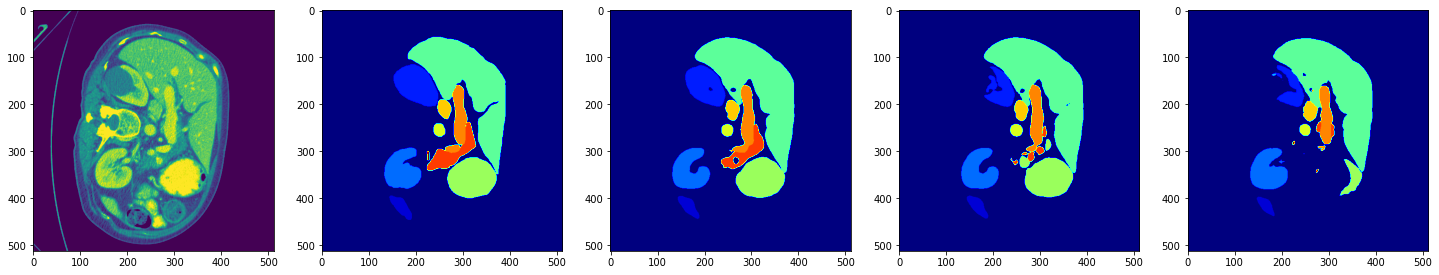}
    \end{minipage} \\
    \begin{minipage}[htbp]{0.75\linewidth}
        \centering
        \includegraphics[width=\linewidth]{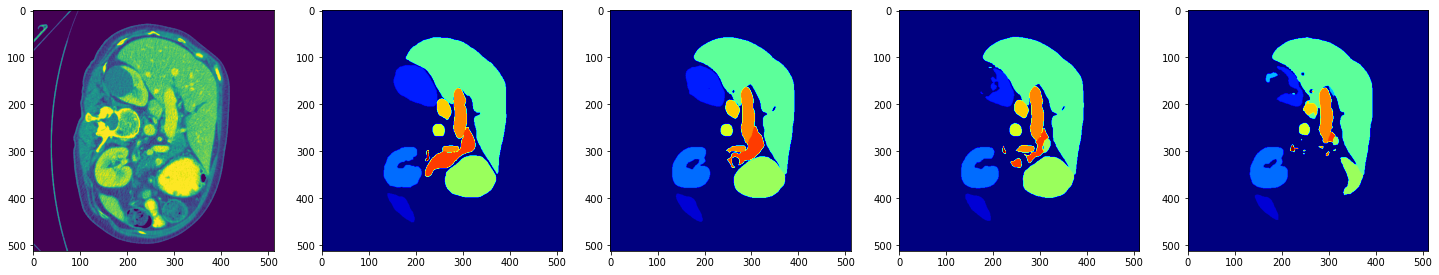}
    \end{minipage} \\
    \begin{minipage}[htbp]{0.45\linewidth}
        \centering
        \includegraphics[width=\linewidth]{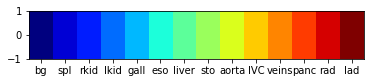}
    \end{minipage}
    \caption{
    \emph{Transfer learning reduces noise in local predictions, and depth information further improves prediction consistency across slices.} We show five consecutive input images, ground-truth labels, predictions from \textit{Ours}, \textit{Ours w/o D}, \textit{Ours w/o T\&D}.
    }
    \label{fig:cross_slice_consistency}
\end{figure*}

Most existing works such as TransUNet~\citep{chen2021transunet} and SwinUNet~\citep{cao2021swin} that use Transformers for 3D medical image segmentation are slice-based approaches, which split a 3D image into 2D slices and predict segmentation independently on each slice. This approach enables the direct utilization of pre-trained weights for transfer learning, but none of these works study the effect of transfer learning by comparing their methods with randomly initialized encoders. We find that \emph{transfer learning significantly boosts the segmentation performance even under significant domain shifts}. Using the pre-trained encoder leads to 11.18\% improvements in DSC as compared to the same randomly initialized encoder (Table~\ref{tab:bcv_seg}(a))\footnote{SwinUNet$_\downarrow$ indicates an input resolution of 384, while other models use an input resolution of 512. TransUNet and SwinUNet results are from their original papers, while other results are from our experiments.}. 

Recent works such as CoTr~\citep{xie2021cotr} or UNETR~\citep{hatamizadeh2021unetr} modify the Transformer architecture to support 3D inputs to leverage depth information. However, they all use randomly initialized 3D Transformers as encoders, which sacrifices the clear benefit of transfer learning as discussed above. Due to the different amount of training data and number of labels reported in their papers\footnote{CoTr use 21 training data from the BCV dataset and reported DSC based on 11 organs, while UNETR uses 28 + external training data and the result is based on an ensemble of four models.}, we reproduced UNETR using its officially released codes for a fair comparison. From Table~\ref{tab:bcv_seg}, we can see UNETR underperforms TransUNet or SwinUNet, but it achieves 6.07\% improvements in DSC compared to the randomly initialized slice-based model, demonstrating \emph{the importance of depth information}.

To combine the advantages of \emph{transfer learning} and \emph{depth information}, we use the weight inflation strategy introduced in Sec.~\ref{section:transfer} to initialize 3D Transformers. We observe consistent improvements with this initialization over random, but we find that the weight inflation depth should not be too large (Sec.~\ref{section:inflation-ablation}).
Our best model uses a window-based approach. Specifically, we split 3D images into small windows along the depth axis, where each window consists of a few (e.g., 5) neighbor slices. We aggregate all the window predictions to form the final prediction. \emph{By incorporating shallow depth information, our method achieves 1.95\% improvements in DSC compared to slice-based methods}. 
More importantly, \emph{such improvement is at only 0.75\% increased computational cost} (213.3 GFLOPS before adaptation vs. 215.0 after). 

We also evaluate our method stability by running our model and baselines on the BCV dataset three times with three random seeds (1234, 5678, 910). Our model (i.e., Ours in Table~\ref{tab:bcv_seg}(a)) achieves 86.65 $\pm$ 0.35 (87.13, 86.52, 86.29 for three runs) DSC, while the model without weight inflation (i.e., Ours w/o D in Table~\ref{tab:bcv_seg}(a)) only achieves 85.14 $\pm$ 0.20 (85.18, 84.88, 85.36 for three runs) DSC. The results indicate \emph{a 98\% level of significance} using the paired t-test.

To better evaluate the improvements of our method, we extend Table~\ref{tab:bcv_seg} and show the DSC results for each of the 8 organs in Table~\ref{tab:organ_analysis}. We find that the most significant improvements from weight inflation are Gallbladder, Pancreas, and Stomach segmentation. We compute different statistics of each organ to understand the reason, including the number of voxels, the number of occurring slices, and the variation between slices (measured by the average DSC between labels of neighbor slices). We find that \emph{the larger the variation between neighbor slices, the more improvements from incorporating depth information through weight inflation}. 
This can be interpreted as \emph{inter-slice prediction consistencies are significantly improved with depth information} based on the qualitative result shown in Figure~\ref{fig:cross_slice_consistency}.

\begin{table*}[t]
\centering
\small
\caption{\emph{Best practice of weight inflation on the BCV dataset.} Initializing the model with the centering inflation strategy and only predicting the center slice's segmentation from a small number of neighbor slices leads to the best performance. }
\label{tab:ablations}
\subtable[
\textbf{Prediction Target}
\label{tab:decoder_depth}
]{
\centering
\begin{minipage}{0.4\linewidth}{\begin{center}
\begin{tabular}{lcc}
\toprule
Prediction & All & \textbf{Center} \\
\midrule
DSC & 73.93 & \textbf{76.33}\\
\bottomrule
\end{tabular}
\end{center}}\end{minipage}
}
\subtable[
\textbf{Weight Initialization}
\label{tab:decoder_width}
]{
\begin{minipage}{0.55\linewidth}{\begin{center}
\begin{tabular}{lccc}
\toprule
Initialization & Random & Average & \textbf{Centering} \\
\midrule
DSC & 72.49 & 76.33 & \textbf{84.30}\\
\bottomrule
\end{tabular}
\end{center}}\end{minipage}
}
\\
\subtable[
\textbf{Input Stride}
]{
\begin{minipage}{0.4\linewidth}{\begin{center}
\begin{tabular}{lccc}
\toprule
Input Stride & \textbf{1} & 2  & 3  \\
\midrule
DSC & \textbf{84.74} & 84.30 & 77.68 \\
\bottomrule
\end{tabular}
\end{center}}\end{minipage}
}
\subtable[
\textbf{Input Slices}
\label{tab:mask_token}
]{
\begin{minipage}{0.55\linewidth}{\begin{center}
\begin{tabular}{lcccc}
\toprule
Input Slices & 1 & \textbf{5} & 11 & 21 \\
\midrule
DSC & 83.34 & \textbf{84.74} & 83.83 & 77.52 \\
\bottomrule
\end{tabular}
\end{center}}\end{minipage}
}
\end{table*}

\subsection{Pre-training Source and Objective} 
\label{section:encoder-ablation}

Since pre-training sources may significantly impact the transfer performance, we compare the transfer effectiveness of various vision Transformers pre-trained with different sources (natural images, medical images, natural videos) and objectives (supervised learning, self-supervised learning, combination of both) (Table~\ref{tab:bcv_seg}(b)). 
To fairly compare models, we carefully select same-size models with similar computations.
SwinT~\citep{liu2021swin} is pre-trained on natural image dataset ImageNet-22K~\citep{deng2009imagenet} via supervised learning.
DINO~\citep{caron2021emerging} is pre-trained on ImageNet-1K via self-supervised contrastive learning.
BEiT-SSL~\citep{bao2021beit} is pre-trained on ImageNet-22K via self-supervised masked image modeling, and BEiT is further fine-tuned on the same dataset via supervised learning. 
VideoSwinT~\citep{liu2021video} is pre-trained on natural video dataset Kinetics-600~\citep{carreira2018short}.
We also pre-trained a CT-DINO on the medical image dataset RSNA-CT~\citep{colak2021rsna} using the same model and objective as DINO. 

Interestingly, we find that \emph{models pre-trained on medical images performs worse than models pre-trained on natural images if both models are pre-trained using self-supervised learning}. This might be attributed to the high inter-class similarity for medical images due to the nature of human anatomy, causing limitations in learning meaningful representations using self-supervised learning. Furthermore, medical image datasets are much smaller than natural image datasets. Designing effective self-supervised learning methods in medical images will be meaningful future work. Moreover, we find that \emph{additional supervision from image labels or video labels further improved the transfer performance}, as models pre-trained via supervised learning generally achieved better performances than models pre-trained via self-supervised learning.

Since our goal is to explore the best strategy to leverage existing large-scale pre-trained 2D vision Transformers given their strong adaptability to various tasks and public availability, we did not perform additional pre-training. Thoroughly exploring pre-training on different sources with different objectives and data amounts is a meaningful future direction but requires extensive computational resources and costs. 

\begin{table*}[t]
\centering
\scriptsize
\caption{\emph{Generalizability of best practice.} To verify the generalization of our best practice that adapts Transformers from 2D to 3D, we compare our adapted Transformers with original Transformers on 11 additional datasets and observe improvements over 10 datasets. DSC excluding background on the test set is reported. We also include other baseline performances for reference, but numbers are not fully comparable due to no standardized data split available for these datasets ({+ indicates same data split ratio but not same split; * indicates different data split ratios}).
}
\begin{tabular}{lccccccccccc}
\toprule
\textbf{Method} & \textbf{ACDC} & \textbf{Bra} & \textbf{Hea} & \textbf{Liv} & \textbf{Hip} & \textbf{Pro} & \textbf{Lun} & \textbf{Pan} & \textbf{Ves} & \textbf{Spl} & \textbf{Col} \\
\midrule
SwinUNet$^+$~(\citeyear{cao2021swin}) & 90.00 & - & - & - & - & - & - & - & - & - & - \\
UNETR$^+$~(\citeyear{hatamizadeh2021unetr})  & - & 71.1 & - & - & - & - & - & - & - & 96.4 & - \\ 
nnUNet$^*$~(\citeyear{isensee2021nnu}) & 91.79 & 74.11 & 93.28 & 79.71 & 88.91 & 75.37 & 72.11 & 67.45 & 68.37 & 96.38 & 45.53 \\
\midrule
Ours w/o D & 92.62 & 79.30 & 91.22 & 87.24 & 86.24 & 74.29 & 62.01 & 58.60 & 67.95 & 95.12 & 46.28 \\
Ours & 92.69 & 80.13 & 90.78 & 87.67 & 86.83 & 75.20 & 70.20 & 59.94 & 70.42 & 95.56 & 51.70  \\
\bottomrule
\end{tabular}
\label{tab:other-dataset}
\end{table*}

\subsection{Best Practice of Weight Inflation}
\label{section:inflation-ablation}

We conduct ablations on different inflation settings, including prediction target, weight initialization, stride, and the number of input slices (Table~\ref{tab:ablations})\footnote{Here DSC is averaged on all the 14 labels instead of 8 major organs in Table~\ref{tab:bcv_seg}.}.
We find that \emph{initializing the model with the centering inflation strategy and only predicting the center slice's segmentation leads to the best performance}. This setting is similar to residual connection~\citep{he2016identity}, where the model gradually learns to utilize additional depth information, and it should not be worse than the model without depth information.
Moreover, we find that \emph{including a few neighbor slices is enough}, because neighbor slices are more similar to the center slice than non-neighbor slices are and provide more information for the model to predict accurate segmentation. Including too many slices may lead to overfitting. This finding also applies to VideoSwinT, where we observe 81.10 and 81.83 DSC for 9 and 5 slices, respectively. 

\subsection{Generalization of Best Practice}
\label{section:practices}

To verify the generalizability of our best practice --- adapting pre-trained vision Transformers from 2D to 3D using the strategies in Table~\ref{tab:ablations}, we train and evaluate our method on 11 additional 3D medical image datasets of different anatomical regions and imaging modalities. 
From Table~\ref{tab:other-dataset}, we observe that \emph{weight-inflated vision Transformers (Ours) achieve consistently better performances on 10 of 11 datasets compared to the original Transformers (Ours w/o D)}, which \emph{clearly demonstrates the effectiveness of our proposed method}. 
Also, these results are obtained with just a single set of hyperparameters tuned on the BCV dataset, \emph{showing the robustness of our method}.

Since our model uses all the hyperparameters tuned on the BCV dataset, continuing tuning hyperparameters on each dataset should lead to further improvements. For example, since the hyperparameters of our method are tuned on the CT dataset, the optimal hyperparameters may be very different for MRI datasets, thus MRI datasets such as Bra, Hea, Hip, and Pro show smaller gains than other CT datasets. However, since the main purpose of our paper is to show the importance of weight inflation, not to achieve different state-of-the-arts, we leave the relation between data characteristics and optimal hyperparameters to future work.

\section{Conclusion}

In this work, we investigated adapting pre-trained Transformers for 3D medical image segmentation via simple yet effective weight inflation strategies. Our approach achieved consistent improvements on 12 datasets with only 0.75\% increased computational cost, which can become a standard strategy easily utilized by all work on Transformer-based models for 3D medical images, to maximize performance.

\acks{
We thank all the reviewers for their constructive feedback. This work is partially supported by the Stanford Center for Artificial Intelligence in Medicine \& Imaging (AIMI).
}

\bibliography{zhang22}

\appendix

\begin{table*}[!t]
\centering
\small
\caption{\emph{Details of the experimental settings}. We do not perform heavy hyperparameter tuning to ensure the generalizability of our best practices. The ensemble is not used, which can further improve performances. *For Brain, Vessel, and Pancreas, we train each model with 250,000 steps given their larger training set. We provide all the codes, including data pre-processing, data loading, model training, and evaluation at \url{https://github.com/yuhui-zh15/TransSeg} } 
\begin{tabular}{llll}
\toprule
\textbf{Hyperparam} & \quad\quad\quad \textbf{Value} & \quad\quad\quad \textbf{Hyperparam} & \quad\quad\quad \textbf{Value} \\
\midrule
Batch Size & \quad\quad\quad 16 & \quad\quad\quad Patch Size & \quad\quad\quad 16$\times$16$\times$5 \\
Loss Function & \quad\quad\quad DiceFocal & \quad\quad\quad Optimizer & \quad\quad\quad AdamW \\
Learning Rate & \quad\quad\quad 3e-5 & \quad\quad\quad Weight Decay & \quad\quad\quad 5e-2 \\
Scheduler & \quad\quad\quad Slanted Triangular & \quad\quad\quad Warm-up Steps & \quad\quad\quad 20 \\
Step & \quad\quad\quad 25,000* & \quad\quad\quad Hyperparam Tuning & \quad\quad\quad No \\
GPUs & \quad\quad\quad 8 Titan RTX & \quad\quad\quad Time & \quad\quad\quad 8 Hours \\
Ensemble & \multicolumn{3}{l}{\quad\quad\quad No multi-model, multi-view, multi-scale ensemble} \\
\midrule
Train Data & \multicolumn{3}{l}{Random Zoom ($[0.5\times, 2\times]$), Random Crop (if Zoom $>1\times$),} \\
& \multicolumn{3}{l}{Normalize Intensity ($[-175, 250]$ (CT) or $[0, \text{MAX}]$ (MRI) $\rightarrow$ $[-1, 1]$),} \\
& \multicolumn{3}{l}{Random Flip ($p=0.1$), Random Rotation ($p=0.1$),} \\
& \multicolumn{3}{l}{Random Shift Intensity ($[-0.1,+0.1]$ with $p=0.5$), Pad (if Zoom $<1\times$)} \\
Inference Data & \multicolumn{3}{l}{Normalize Intensity (same as training)} \\
\bottomrule
\end{tabular}
\label{tab:experimental-details}
\end{table*}

\end{document}